\documentclass{article}

\usepackage{arxiv}

\usepackage[utf8]{inputenc} 
\usepackage[T1]{fontenc}    
\usepackage{hyperref}       
\usepackage{url}            
\usepackage{booktabs}       
\usepackage{amsfonts}       
\usepackage{nicefrac}       
\usepackage{microtype}      
\usepackage{lipsum}
\usepackage{graphicx}
\graphicspath{ {./images/} }

\title{Rice Leaf Disease Classification and Detection Using YOLOv5}

\author{
 Md Ershadul Haque \\
 School of Computing, Mathematics and Engineering\\
  Charles Sturt University \\
 Bathurst, Australia \\
  \texttt{mhaque@csu.edu.auu} \\
   \And
 Ashikur Rahman \\
  Department of Electrical and Electronic Engineering\\
  Feni University\\
  Feni, Banglades \\
  \texttt{ashik.fu.eee.02@gmail.com} \\
  \And
  Iftekhar Junaeid \\
  Department of Electrical and Electronic Engineering\\
  Feni University\\
  Feni, Banglades \\
  \texttt{ifte06feni.uv@gmail.com} \\
  \And
  Samiul Ul Hoque \\
  Department of Electrical and Electronic Engineering\\
  Feni University\\
  Feni, Banglades \\
  \texttt{samiul0906@gmail.com} \\
  \And
  Manoranjan Paul \\
 School of Computing, Mathematics and Engineering\\
  Charles Sturt University \\
 Bathurst, Australia \\
  \texttt{mpaul@csu.edu.auu} \\
}

\begin{document}
\maketitle
\begin{abstract}
A staple food in more than a hundred nations worldwide is rice (Oryza sativa). The cultivation of rice is vital to global economic growth. However, the main issue facing the agricultural industry is rice leaf disease. The quality and quantity of the crops have declined, and this is the main cause. As farmers in any country do not have much knowledge about rice leaf disease, they cannot diagnose rice leaf disease properly. That's why they cannot take proper care of rice leaves. As a result, the production is decreasing. From literature survey, it has seen that YOLOv5 exhibit the better result compare to others deep learning method. As a result of the continual advancement of object detection technology, YOLO family algorithms, which have extraordinarily high precision and better speed  have been used in various scene recognition tasks to build rice leaf disease monitoring systems. We have annotate 1500 collected data sets and propose a rice leaf disease classification and detection method based on YOLOv5 deep learning. We then trained and evaluated the YOLOv5 model. The simulation outcomes show improved object detection result for the augmented YOLOv5 network proposed in this article. The required levels of recognition precision, recall, mAP value, and F1 score are  90\%, 67\%, 76\%, and 81\%  respectively are considered as performance metrics. 
\end{abstract}


\section{Introduction}
The production of rice is recognized as the single most significant economic activity on the globe since it is the commercially and culturally most significant food crop. There are more than 2.7 billion people on the planet, the majority of whom consume rice as their main meal. This estimate will increase to 3.9 billion people by the year 2025. One of the top five producers and consumers of rice worldwide is Bangladesh. About 135 million people in Bangladesh, an agricultural country, choose rice as their primary food \cite{b1}. During the most recent fiscal year (FY'21), the country's rice ministry produced 37.5 million tonnes, while importing 1.35 million tonnes. Bangladesh has been producing more rice each year, as illustrated in Fig. 1.
\begin{figure}[htbp]
\centerline{\includegraphics[width=\linewidth]{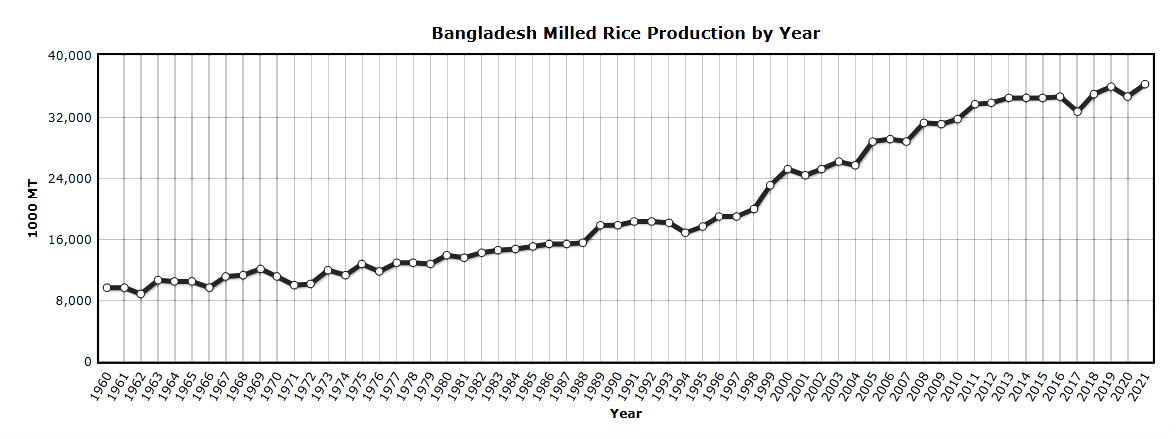}}
\caption{Production of rice Leaf import by countries.}
\label{fig1}
\end{figure}
But diseases of the rice leaves are currently severely impeding rice output. The amount of rice produced in our nation is much less due to the prevalence of rice leaf diseases. According to \cite{b2}, rice disease damage can considerably reduce productivity. They are primarily caused by bacteria, viruses, or fungi. Applying computer vision technology to the management of diseases is the simplest and frequently most cost-effective technique.
\begin{figure}[htbp]
\centerline{\includegraphics[width=\linewidth]{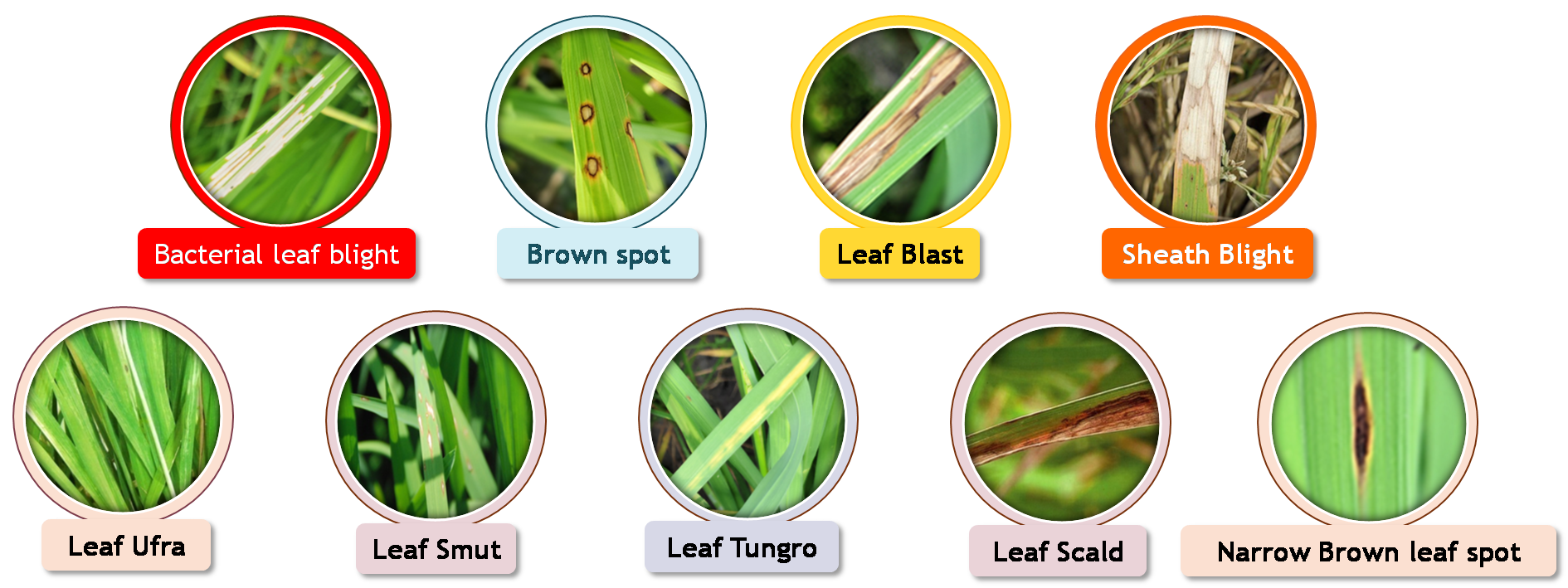}}
\caption{Rice leaf disease classification.}
\label{fig2}
\end{figure}
The global spread of the rice disease is seen in Fig. 2. Bacterial leaf blight, Brown Spot, Leaf Blast, Leaf Scald, Leaf Tungro, Leaf Ufra, Narrow Brown Leaf Spot, and Sheath Blight are a few of the diseases that can affect rice leaves.
\begin{figure}[htbp]
\centerline{\includegraphics[width=\linewidth,height=6cm,width=6cm]{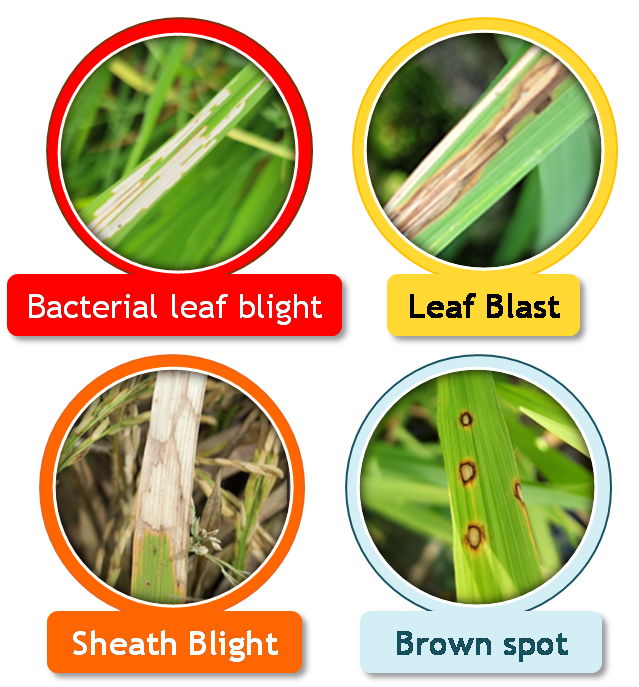}}
\caption{Major Rice leaf disease prospective of Bangladesh .}
\label{fig3}
\end{figure}
Fig. 3 represents the four rice leaf diseases we discovered throughout our research: bacterial leaf blight, brown spot, leaf blast, and sheath blight. In Bangladesh, these illnesses are the most prevalent types of rice leaf disease. Xanthomonas oryzae pv. oryzae is the culprit behind bacterial blight. Seedlings wilt, and leaves turn yellow and dry out as a result \cite{b3}. The coleoptile, leaves, leaf sheath, panicle branches, glumes, and spikelets are all affected by the fungus known as brown spot. The multiple large spots on the leaves, which can kill the entire leaf, are the damage that is easiest to see. Unfilled grains or speckled or discolored seeds develop when the seed becomes infected \cite{b4}.
\begin{figure}[htbp]
\centerline{\includegraphics[width=\linewidth,height=6cm,width=6cm]{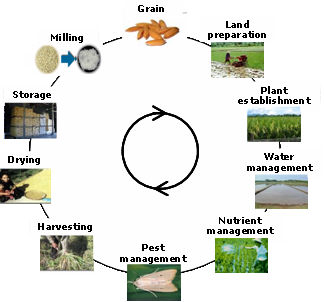}}
\caption{ Rice production step in Bangladesh .}
\label{fig4}
\end{figure}
The fungus Magnaporthe oryzae is responsible for blasting. All of a rice plant's above-ground components, including the leaf, collar, node, neck, portions of the panicle, and occasionally the leaf sheath, are susceptible to it \cite{b5}. Rhizoctonia solani is a fungus that causes the disease known as sheath blight. Leaves that are infected senesce or dry out and die more quickly. You can also eliminate young tillers\cite{b6}. The reliable diagnosis of rice leaf disease is presently made possible by convolutional neural network (CNN) technology. CNN is a part of an artificial neural network. As a consequence, we are able to operate with enormous data. The transfer learning approach mainly deals with vast volumes of data. Machine learning (ML) research on transfer learning (TL) is concerned with the storing of information discovered while addressing one issue and its subsequent application to another problem that is comparable but unrelated.

In this study, we used Yolo v5 of the Convolutional Neural Network, which has been upgraded (CNN) One of the deep learning algorithms, CNN has been gleefully used to tackle computer vision issues including picture classification, object segmentation, image analysis, and more. The object recognition method YOLO (You only live once) separates pictures into a grid structure. Object detection is the responsibility of each grid cell. We thus opted to communicate the most recent findings using YOLOv5. As far as we are aware, this is the first study to employ YOLOv5 to identify illness on rice leaves. Therefore, the objective of this study was to identify the illness from the photographs that were recorded using the YOLOv5.
\section{Objective}
Our key goal in this preparation is to use technology to quickly and accurately classify diseases of rice leaves. As a result, after reading and analyzing the research papers of many authors, we discovered that the Yolov5 Algorithm can quickly and effectively diagnose and classify leaves in big paddy fields, saving time and labor for the farmers. Our main objective is to cultivate paddy with a good yield and without any diseases.
\\
\section{Literature Review}
Inception ResNet V2 is a kind of CNN model. Which is commonly used in the method of transfer learning, for the identification of rce leaf disease.   This model has been optimized for classification work. 95.67\% accuracy was, of this model \cite{b7}.
\\
Through CNN Architecture VGG16, training and testing have been done on paddy fields and data sets collected from the internet. Applying this model,92.42\% accuracy was obtained. Here 1649 rice leaf pictures have been worked on the three main diseases here are bacterial leaf blight, brown spot, and rice leaf blast  \cite{b8}.
\\
The main object of this research paper is to create an object detection application that could detect rice leaf blast and brown spot disease. Research is done on 200 custom data sets, using the YOLO(You Only Look Once) algorithm. Where there was about 90\% accuracy  \cite{b9}.
\\
For the rice leaf disease detection system, artificial intelligence techniques and computer concepts have been used here. Here they found three diseases of rice, Brown spot, Hispa, and leaf blast from 300 pictures. In this study, 97.50\% got accuracy  \cite{b10}
\\
Support vector machines are yet another method of image processing for the identification of rice leaf disease (SVM). Accuracy of 97.2\% has been achieved with this procedure. The infected spot was extracted from the spot using picture, shape, and texture attributes. The SVM approach has been used to identify the causes of bacterial leaf blight, rice sheath blight, and rice blast  \cite{b11}.
\\
Herbal leaves can be classified by a neural network model using YOLO. There are five types of herbal leaves like mehndi, betel leaf, mint, and aloe vera are classified in this model. The accuracy of the classification of this model is about 95\%  \cite{b12}.
\\
In this research paper, yolo v5 is an effective method for detecting masks using a deep learning model. The algorithm is trained in 5 different types of number. By examining 86 images, the deep learning model achieved maximum 96.5\% accuracy through 300 epochs /cite{b13}.
\\
In this paper, research has been done using data sets of 2050 images. Here are 850 pictures of their own, as well as 1200 pictures from the internet. Here they have trained the images using the YOLO V5 algorithm. YOLO V5 accuracy is much higher than YOLO V3 and YOLO V4. Here precision, recall, and average precision (AP) were 98.10\%, 100\%, and 99.60\% respectively  \cite{b14}.
In Fig. 5, we analyze the above mentioned papers and present their used models and accuracy through a graph.
\begin{figure}[htbp]
\centerline{\includegraphics[width=\linewidth]{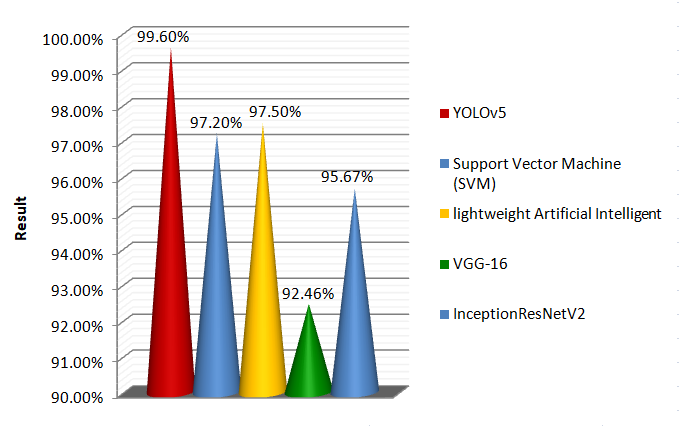}}
\caption{Model Accuracy Graph.}
\label{fig5}
\end{figure}
\\
\section{Materials and methods}
The method of research work preparation consists of six steps: 1) Data collection, 2) Data prepossessing, 3) Data resize, 4) Data annotation, 5) Data augmentation and 6) Data train, as illustrated Figure 6.
\begin{figure}[htbp]
\centerline{\includegraphics[width=\linewidth]{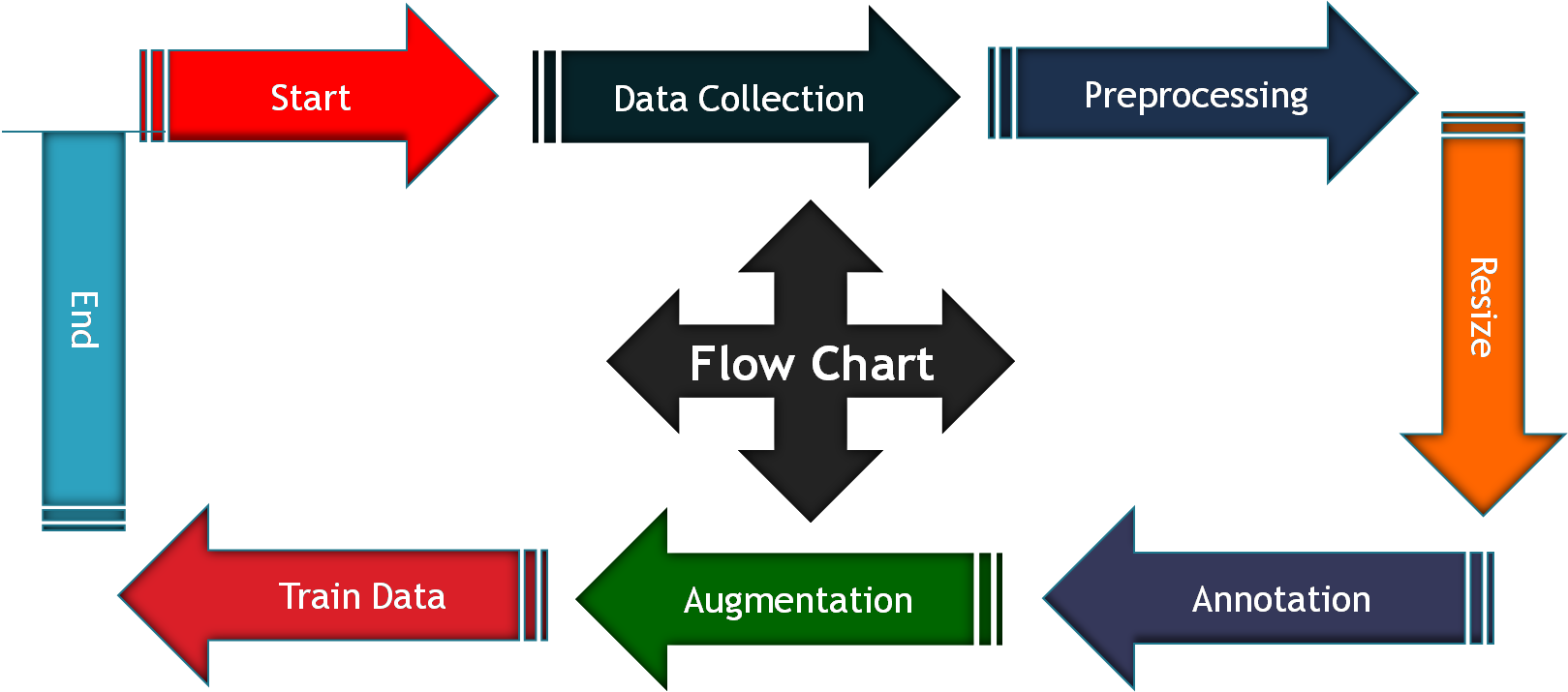}}
\caption{Flow chart of Rice leaf disease.}
\label{fig6}
\end{figure}
\subsection{Data collection}\label{AA}
Images of rice leaves were gathered for the dataset at Feni, Bangladesh. We used a 48MP camera, 4GB RAM, 64GB ROM, and Redmi Node 10 for data collecting. Contains two different kinds of data files Images of health in A; illustrations of diseases in B The dataset includes four different disease categories, as seen in Figure 7. A average of 1500 photos from the bacterial leaf blight, brown spot, leaf blast, sheath blight, and healthy categories were captured.
\begin{figure}[htbp]
\centerline{\includegraphics[width=\linewidth]{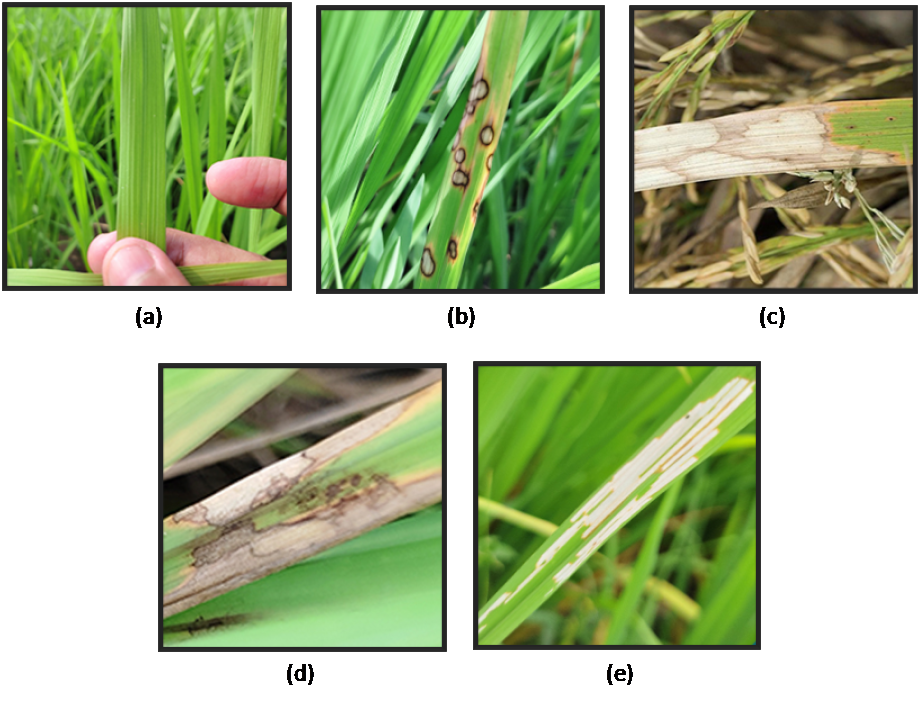}}
\caption{(a) Healthy (b) Brown Spot 
(c) Sheath Blight (d) Leaf  Blast and (e) Bacterial leaf Blight.}
\label{fig7}
\end{figure}
\subsection{Data resize}
We performed picture resizing and changed all of the images in the collected dataset from diverse (height x weight) sizes to 416px to 416px. Photoshop program resizes and uses this dataset. Fig. 8 displays the image of the data resizing.
\begin{figure}[htbp]
\centerline{\includegraphics[width=\linewidth]{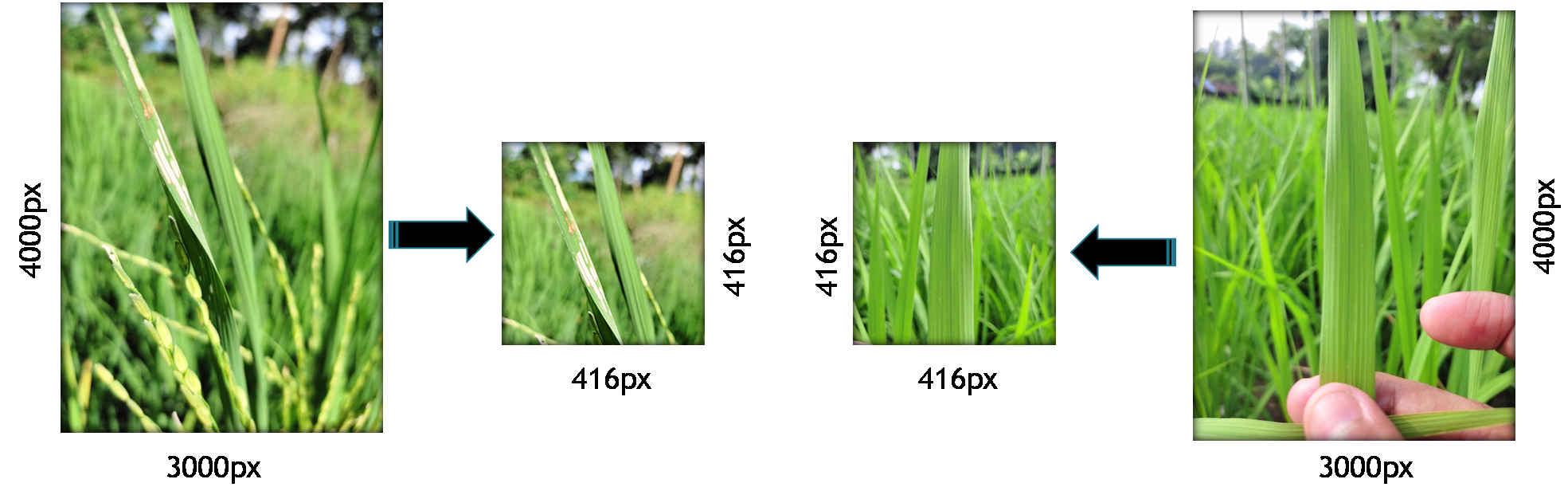}}
\caption{Data Resize}
\label{fig8}
\end{figure}
\subsection{Annotation}
Using Roboflow, a well-known annotation tool, the gathered and resized dataset photographs were cautiously tagged in this final step. This application first opens each image one at a time. Then, to accurately represent an item's position by x, y, and height in that image, a rectangle shape was manually drawn to the edge of the object. Each product's labels list terms like "Healthy," "Brown Spot," "Leaf Blast," "Sheath Blight," and "Bacterial Leaf Blight," among others. Annotated values were saved as YOLOv5 text files during labeling. Data about annotations are shown in Fig. 9.
\begin{figure}[htbp]
\centerline{\includegraphics[width=\linewidth]{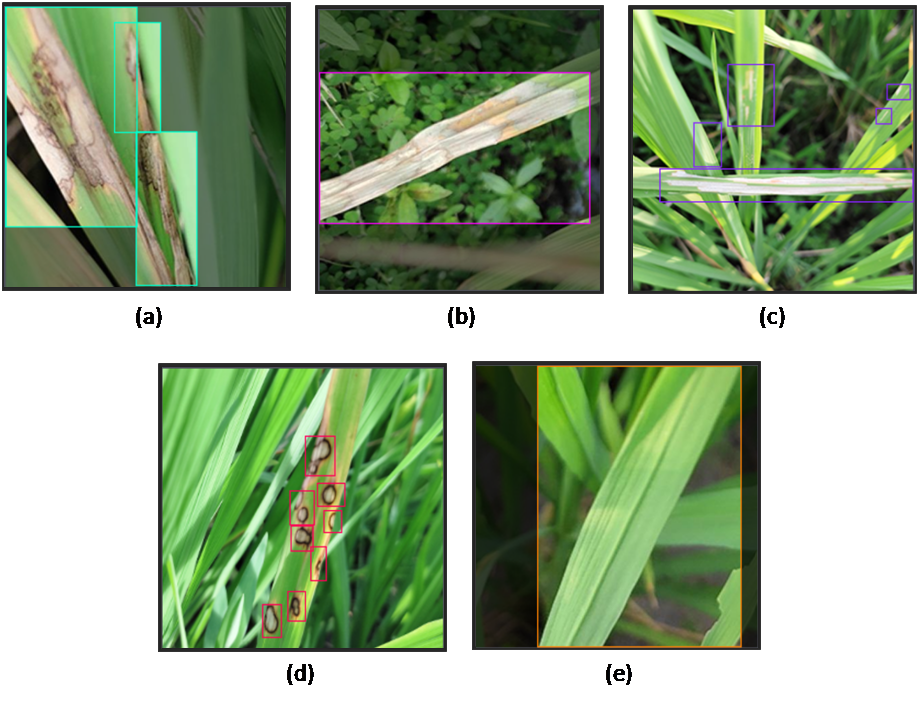}}
\caption{Data Annotation.}
\label{fig9}
\end{figure}
\subsection{Data pre-processing}
These datasets allow for the division of the test set, validation set, and training set into three groups. If 70\% of the images come from the test set, 20\% belong to the validation set, and 10\% are taken from train set.
\subsection{Data augmentation}
The images were labeled and then uploaded to "Roboflow" for augmentation. The goal of data augmentation was to expand the volume and variety of data. It contributed to a decrease in overfitting in small datasets. Fig. 10 illustrates the application of a few data augmentation techniques, including flipping, cropping, and color space change, to create new images from the original dataset.
\begin{figure}[htbp]
\centerline{\includegraphics[width=\linewidth]{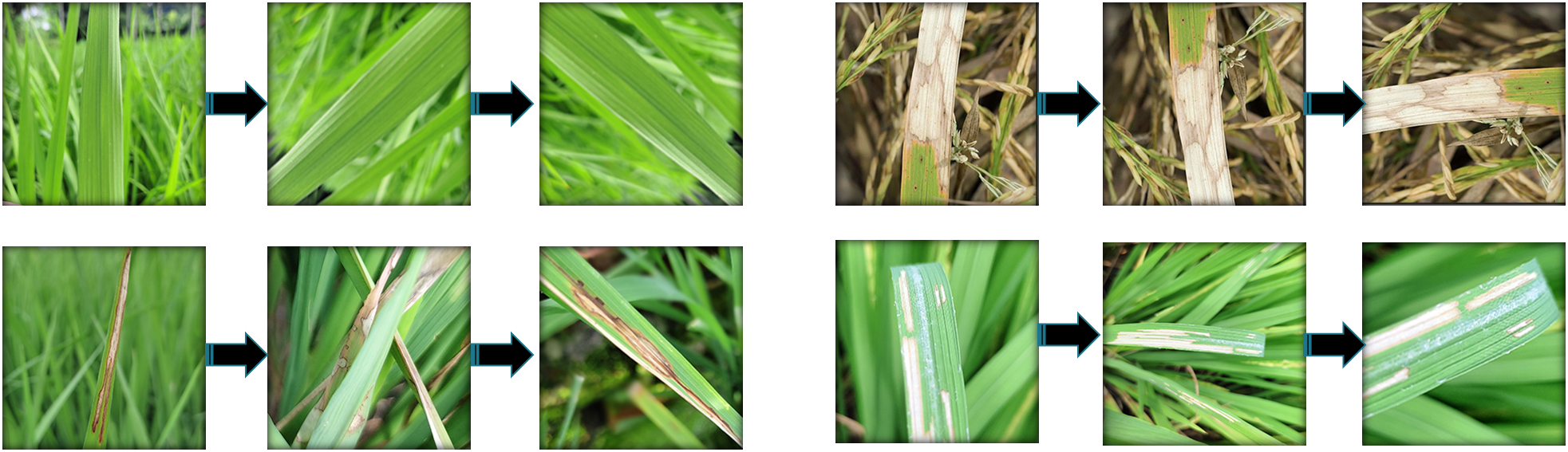}}
\caption{Data Augmentation.}
\label{fig10}
\end{figure}
\begin{figure}[htbp]
\centerline{\includegraphics[width=\linewidth,height=12cm,width=12cm]{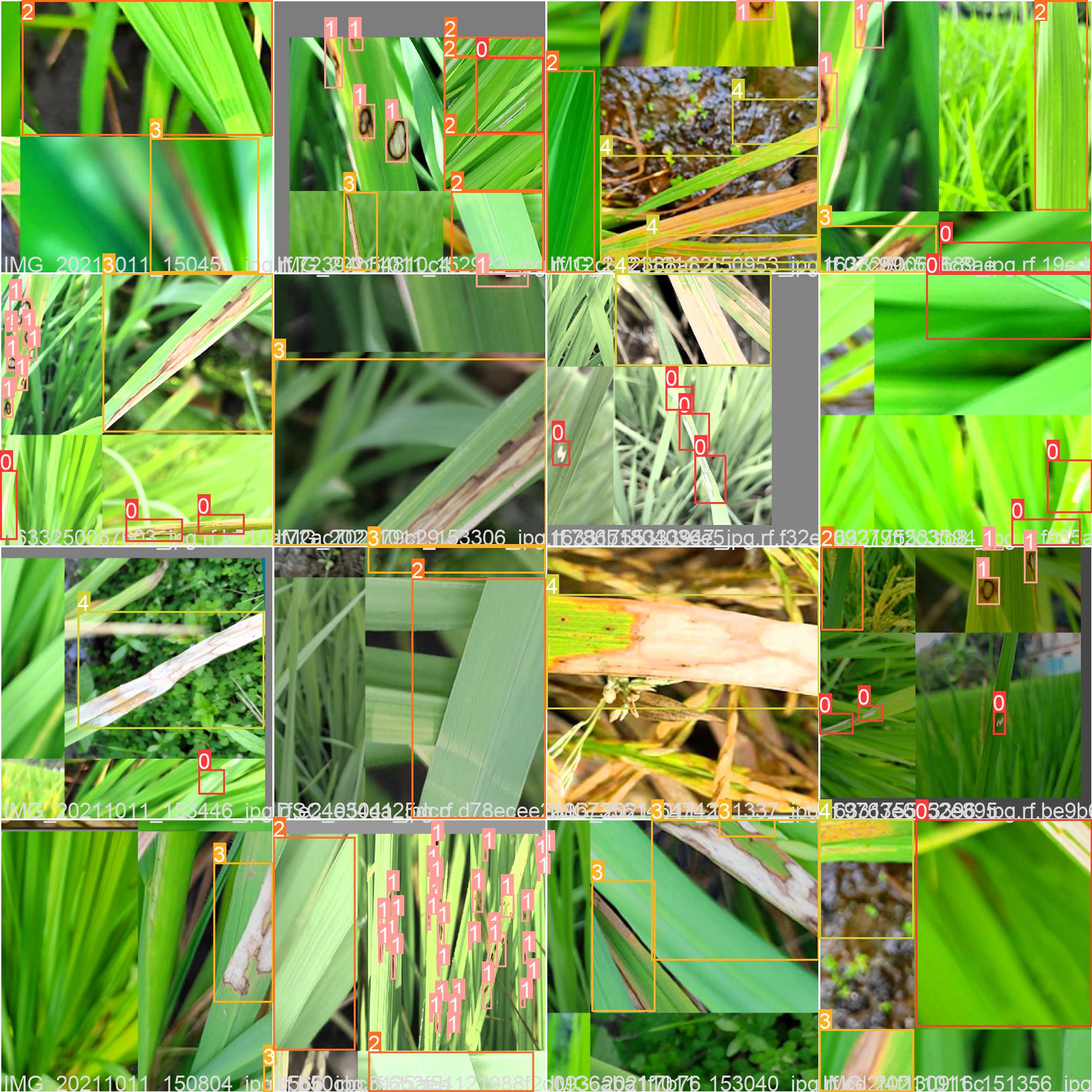}}
\caption{Train batch data.}
\label{fig11}
\end{figure}
\begin{figure}[htbp]
\centerline{\includegraphics[width=\linewidth,height=12cm,width=12cm]{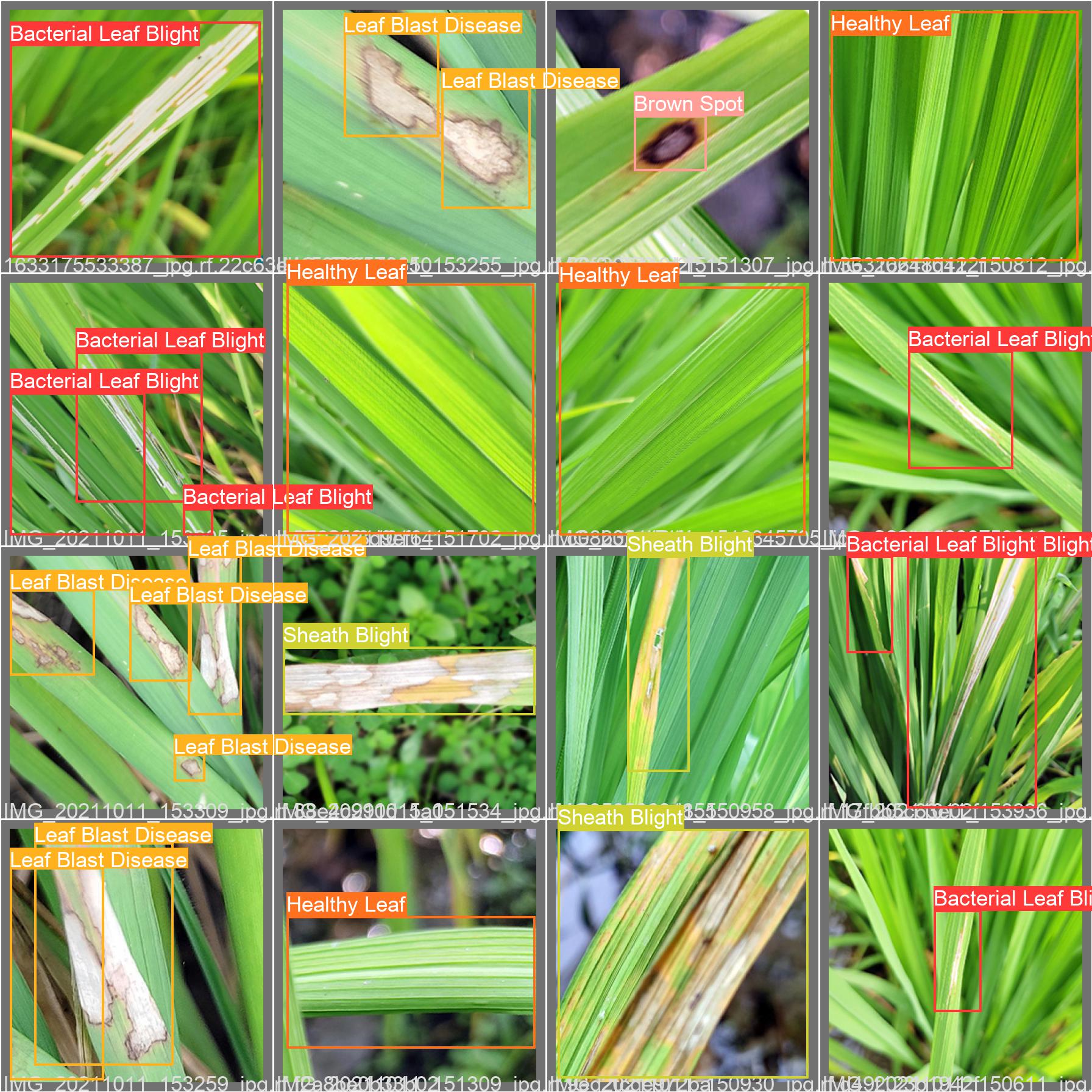}}
\caption{Val batch data.}
\label{fig12}
\end{figure}
\begin{figure}[htbp]
\centerline{\includegraphics[width=\linewidth,height=12cm,width=12cm]{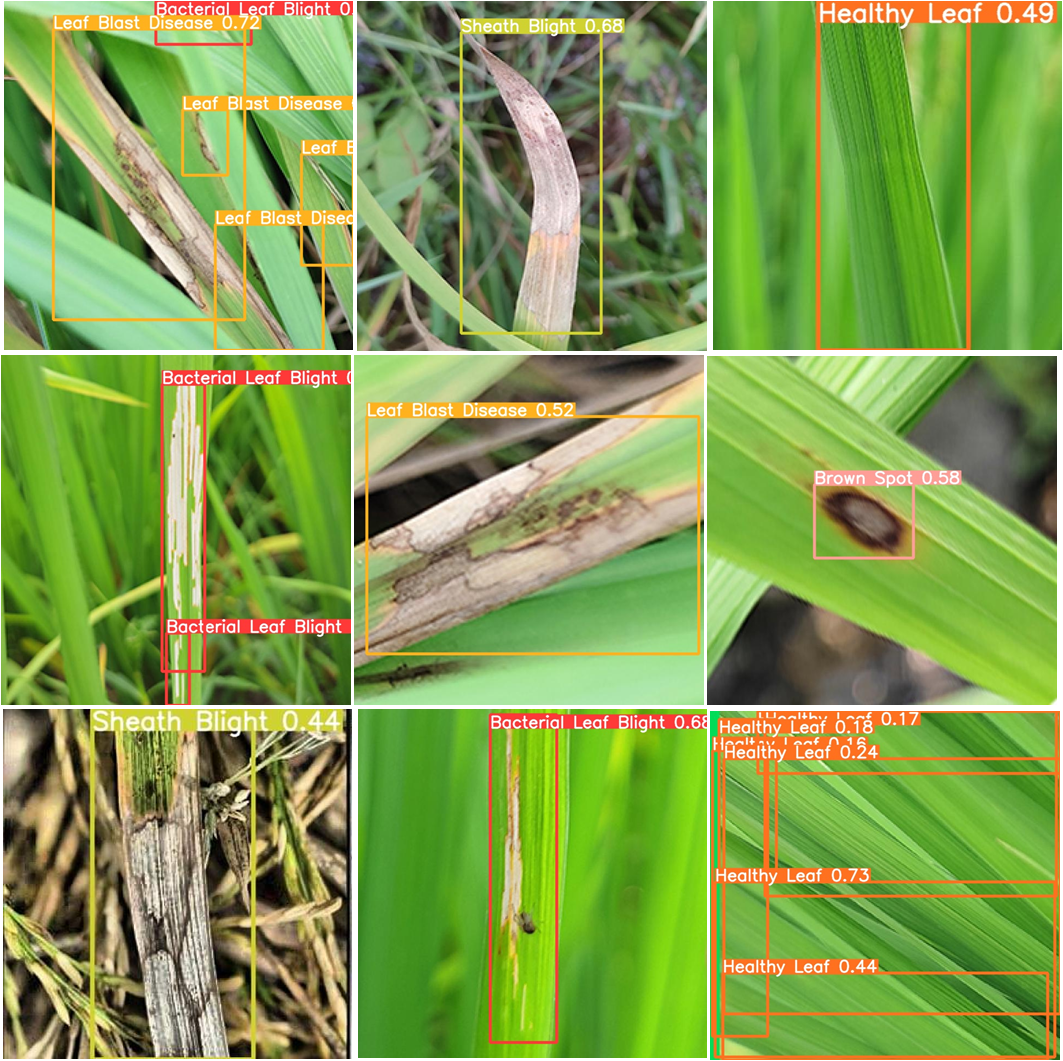}}
\caption{Test batch data.}
\label{fig13}
\end{figure}
\subsection{Proposed method}
We made use of a Roboflow ai notebook (Roboflow, 2016), which is built on YOLOv5 (Jocher et al., 2021) and employs pre-trained COCO weights. From the dataset, Roboflow produced a URL. This URL secures For the objectives of training and assessing the model, Google Colab was utilized. Access to powerful GPUs is free thanks to Google Colab. 150 epochs were chosen, and the training of the model took around 20 minutes. A file named "train.py" may be found in the YOLOv5 directory and is used to train the model. Results from training the YOLOv5 model using photos from the train batch, validation batch, and test batch.
\begin{figure}[htbp]
\centerline{\includegraphics[width=\linewidth]{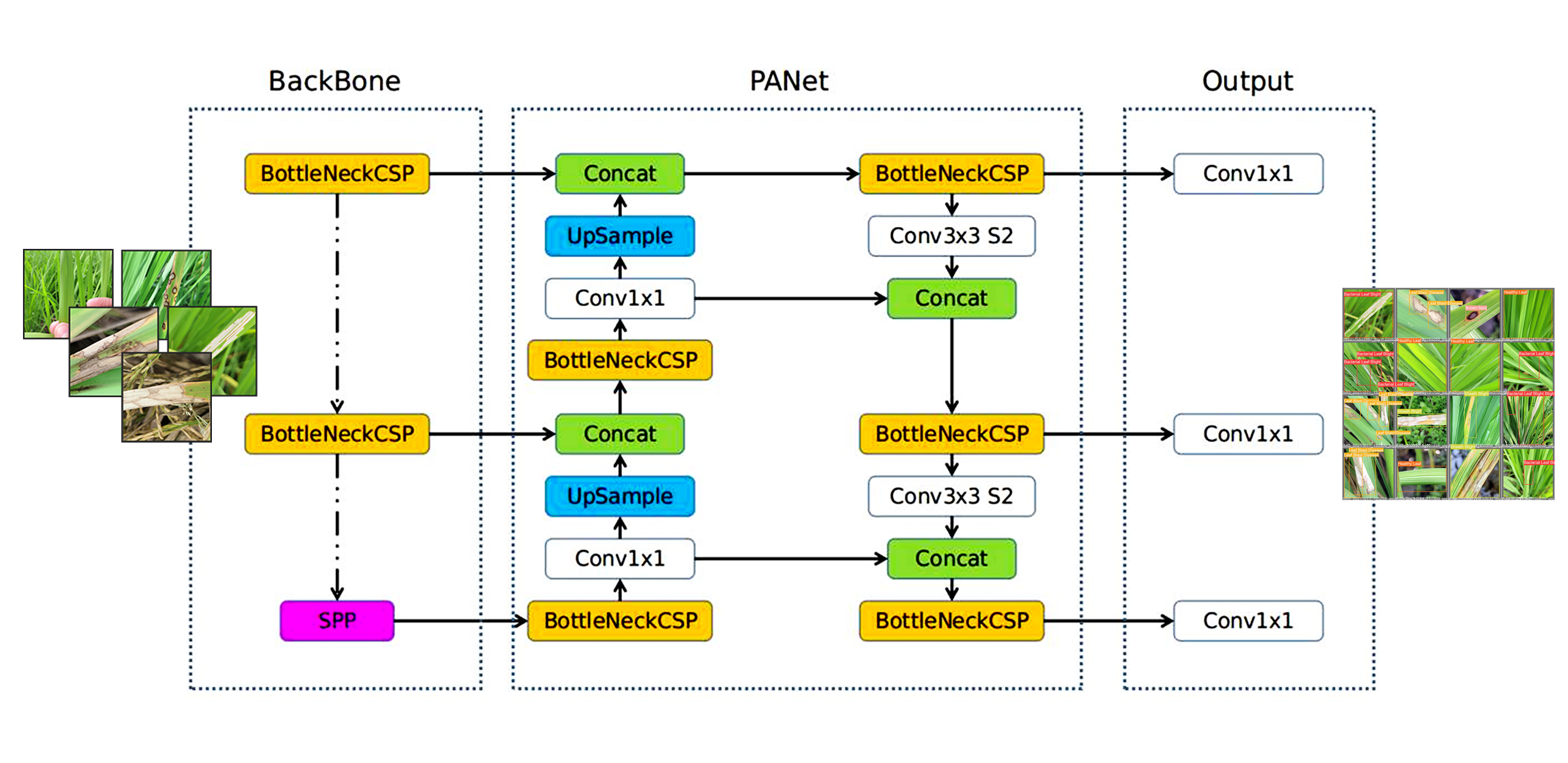}}
\caption{YOLOv5 Architecture}
\label{fig14}
\end{figure}
\subsection{YOLOv5 architecture}
The architecture of Yolov5 is separated into three sections, as shown in Fig. 14: (a) the backbone; (b) the neck; and (c) the output.

1. Backbone: An input image is utilized to identify significant details using the Model Backbone. CSP(Cross Stage Partial Networks) are the framework used in YOLOv5 to extract valuable, useful attributes from an input picture. By conducting feature extraction on the feature map, the BottleneckCSP module primarily obtains rich data from the picture. Compared to other large-scale convolutional neural networks, the BottleneckCSP structure can reduce the duplication of gradient information during the optimization of convolutional neural networks. Its parameter amount makes up the bulk of the network's total parameter quantity. The SPP module adds functionality of various sizes and largely increases the network's receptive area.
2. Neck: Feature pyramids are frequently created by the model's neck. When object scaling is involved, feature pyramids help models generalize to a wider range of situations. It assists in locating an analogous object in different scales and sizes. Other models, including the FPN, BiFPN, and PANet, employ other feature pyramid strategies. PANet is used by YOLO v5 as a neck to obtain a feature pyramid.

3. Head: The head is typically the final stage of the detecting procedure. It creates final output vectors with bounding boxes, objectness scores, and class probabilities using anchor boxes \cite{b15,b16}.
\\
\section{Results and analysis}
According to Equation (3). The model's test accuracy is another factor. The F1 score can range from a minimum of 0, which indicates either zero accuracy or zero recall, to a maximum value of 1, which indicates perfect precision and memory. Additionally, mAP is determined by averaging the average precision (AP) across all classes. Equation (4) illustrates how to calculate mAP by taking the mean of AP. The accuracy of machine learning algorithms can also be determined using mAP. The number of good (safe and uncluttered) landing locations found by the algorithm is the True Positive in the emergency landing spot recognition issue. The number of non-good landing places that the algorithm mistakenly identified as good landing spots is known as the false positive, and the number of good landing spots that the algorithm missed is known as the false negative.
\\
\begin{equation}
Precision =\frac{TP}{(TP + FP)}  
\label{precision}
\end{equation}
\\ 
\begin{equation}
Recall = \frac{TP}{(TP + FN)} 
\label{Recall}
\end{equation}
\\
\begin{equation}
F1 =\frac{2*Precision*recall}{(Precision+Recall)} 
\label{F1}
\end{equation}
\\
\begin{equation}
mAP=\frac{1}{11}*
\sum{r\in(0,0.1,0.2, …..1)}pinterp(r)
\label{AP}
\end{equation}
$$
TP=True Positive, TN=True Negative
$$
$$
FP=False Positive, FN=False Negative.
$$
\\
As a YOLOv5 feature, the model enhances the diagnosis of small-to-large disorders. Precision, recall, F1-score, and mAP of the models all improved. It has also been demonstrated that training performance was more effective with high-quality images.
\begin{figure}[htbp]
\centerline{\includegraphics[width=\linewidth,height=8cm,width=12cm]{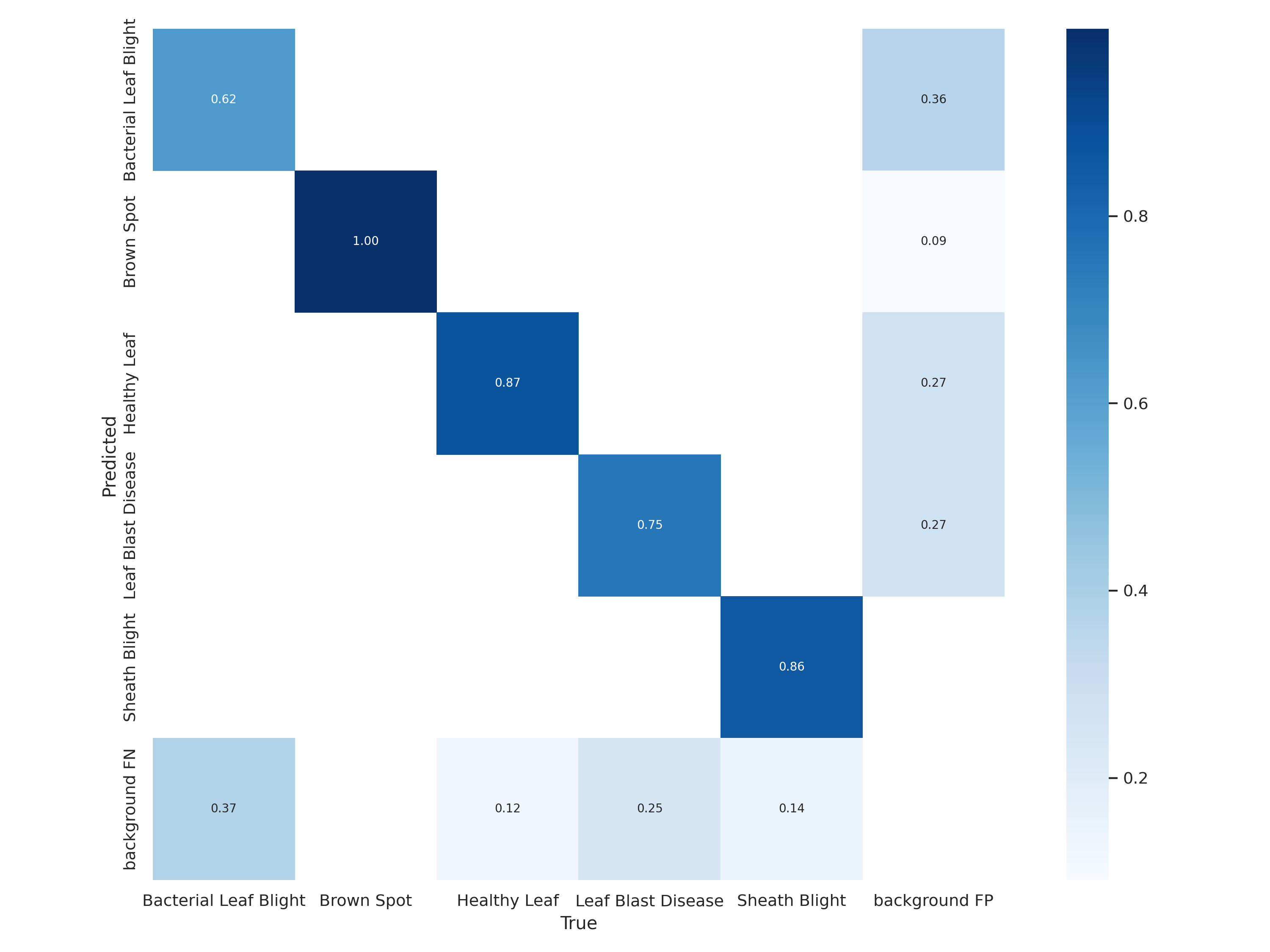}}
\caption{Confusion matrix}
\label{fig15}
\end{figure}
Confusion matrices are used to show crucial predictive data including recall, specificity, accuracy, and precision. Confusion matrices are useful because they offer direct comparisons of variables like True Positives, False Positives, True Negatives, and False Negatives. Fig. 15 shows this.
\begin{figure}[htbp]
\centerline{\includegraphics[width=\linewidth,height=8cm,width=12cm]{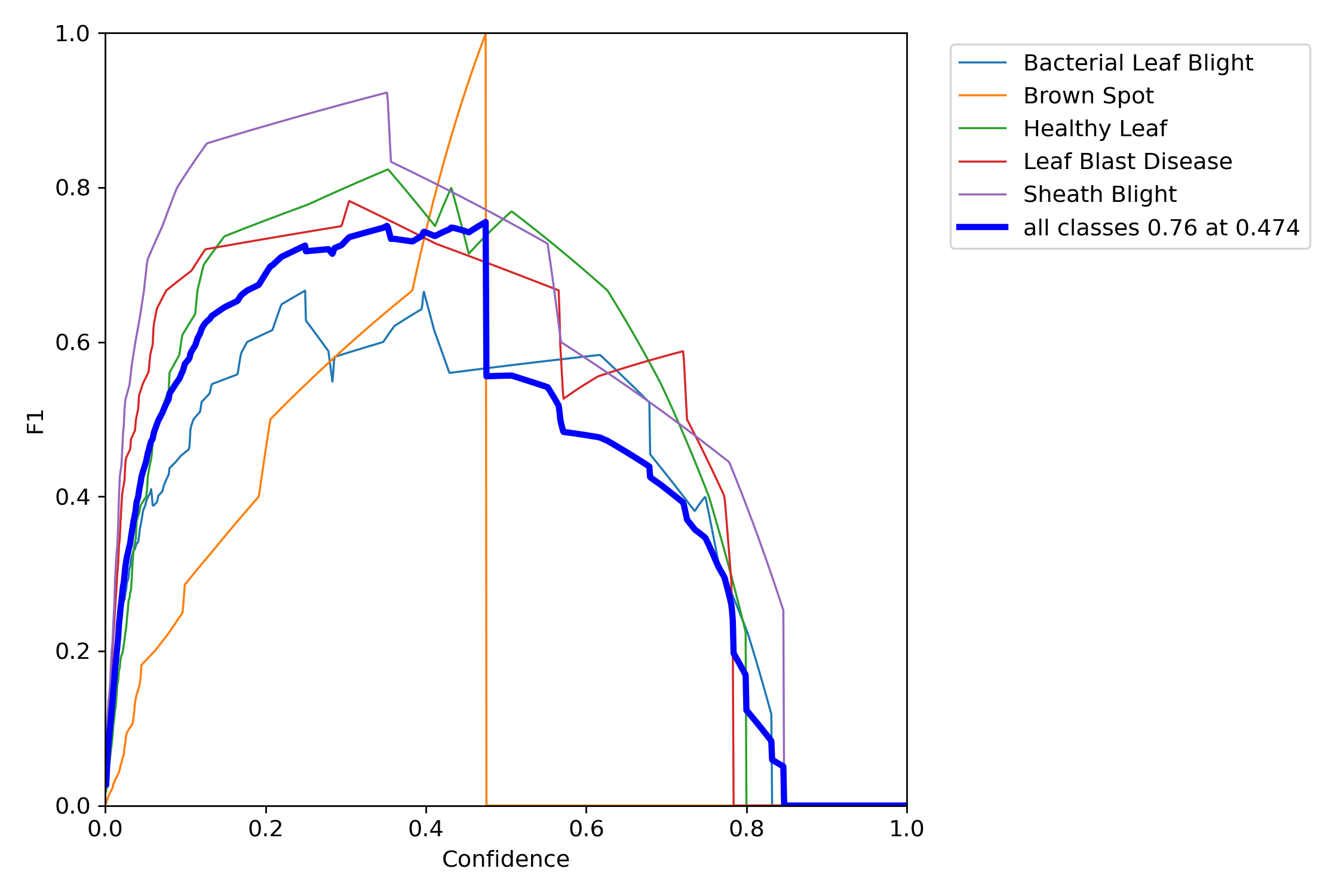}}
\caption{F1 Curve}
\label{fig16}
\end{figure}
The weighted harmonic mean of a classifier's precision (P) and recall (R), using the F1 score, is the F-measure. It implies that both measurements are equally significant. The confidence value in your graph is 0.474, which corresponds to the highest F1 value of 0.76 and optimizes the precision and recall. As demonstrated in Fig. 16, a greater confidence value and F1 score are generally preferred.
\begin{figure}[htbp]
\centerline{\includegraphics[width=\linewidth,height=8cm,width=12cm]{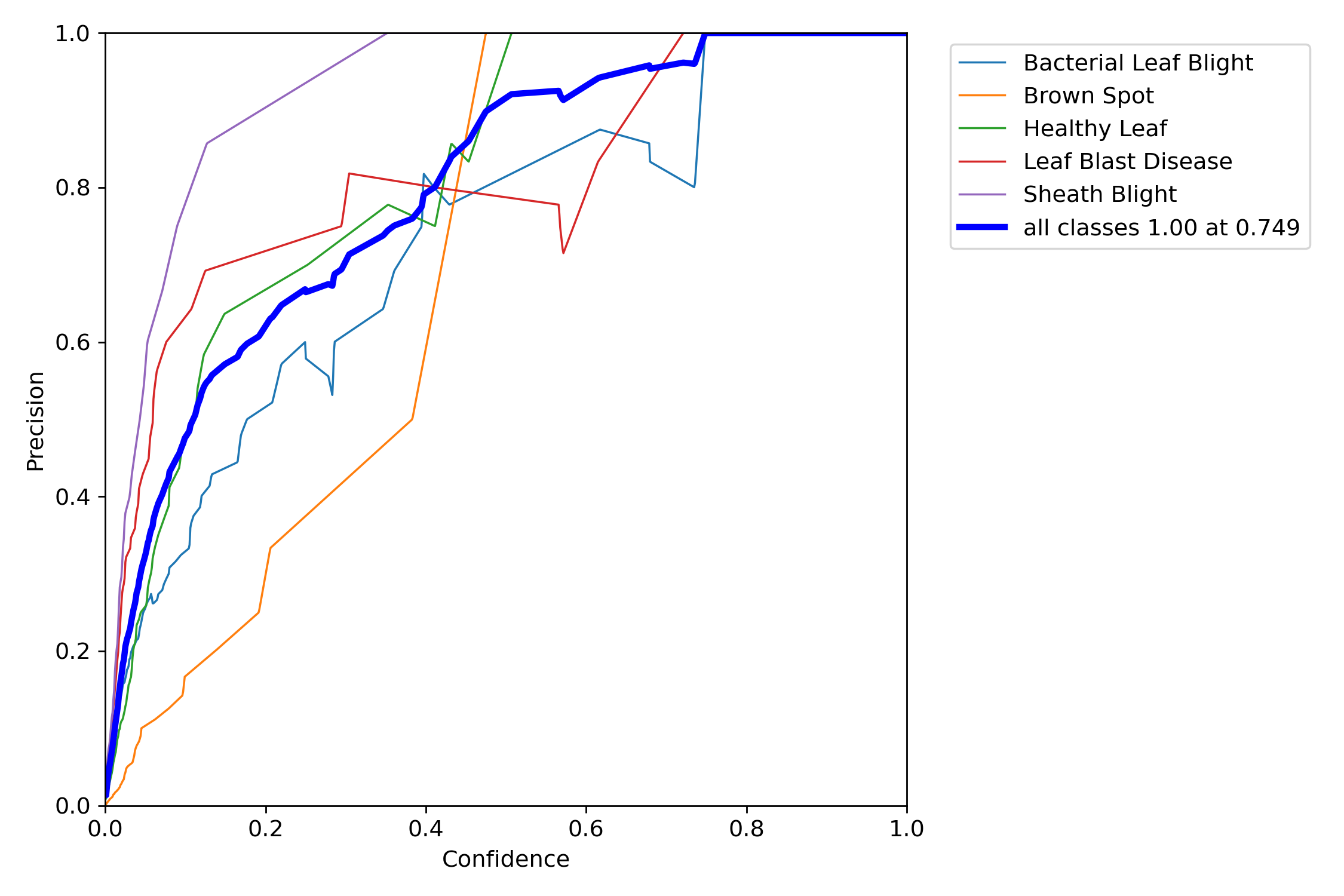}}
\caption{P Curve}
\label{fig17}
\end{figure}
The confidence interval demonstrates the accuracy with which we can define the effect magnitude, and the larger the sample, the more precise the estimate. Practically speaking, the important factor in determining accuracy is typically the sample size. Together, the accuracy value and the confidence interval make sense. Figure 17 demonstrates that the 0.749 confidence range for an effect includes the 1.00 precision values.
\begin{figure}[htbp]
\centerline{\includegraphics[width=\linewidth,height=8cm,width=12cm]{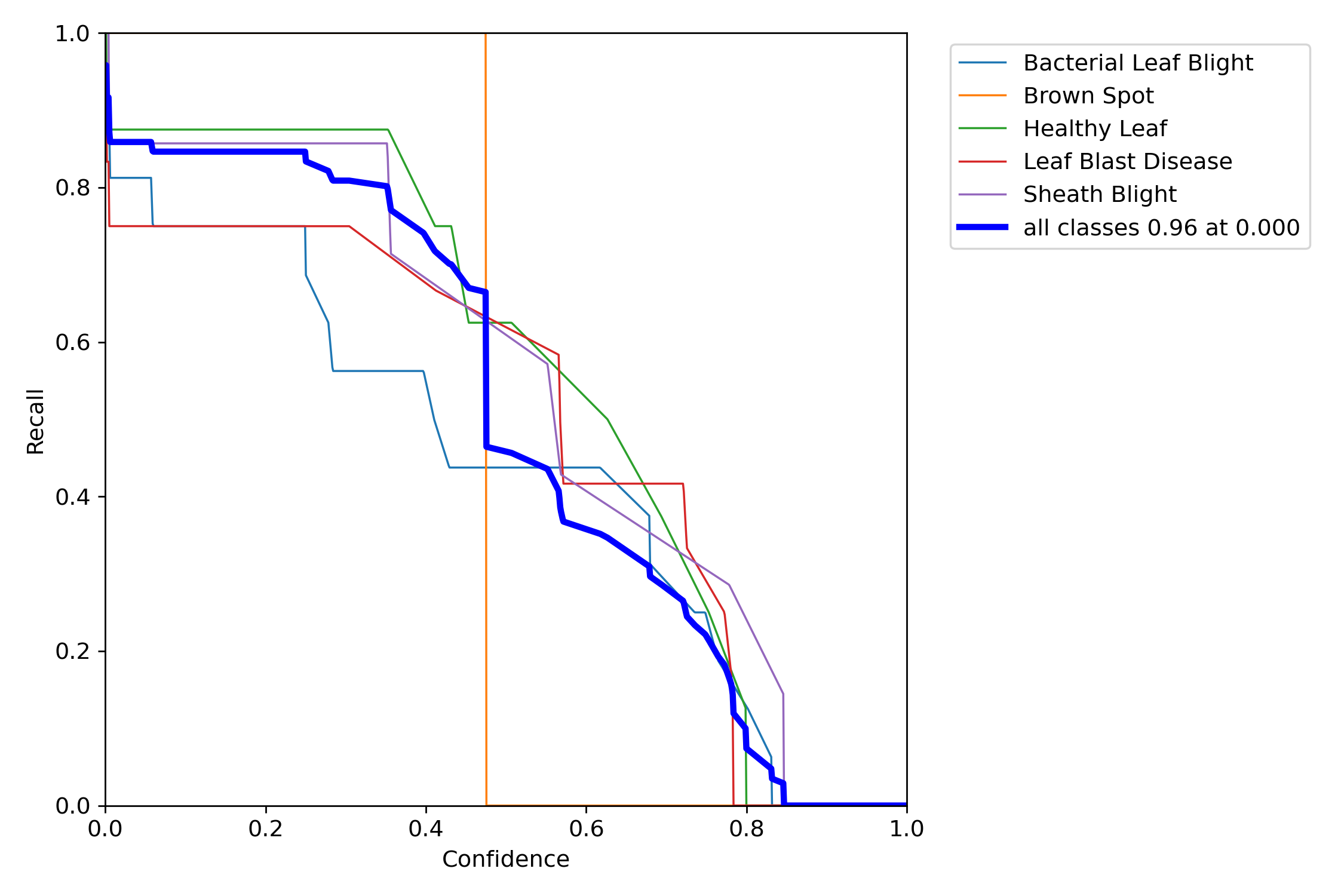}}
\caption{R Curve}
\label{fig18}
\end{figure}
The larger the sample, the more accurate the estimate, and the confidence interval denotes the recall with which we are able to report the effect magnitude. Practically speaking, sample size is frequently the key element in assessing recall. The recall value and confidence interval make logical sense together (see Fig.18). The 0.00 recall values include the effect's 0.96 confidence interval.
\begin{figure}[htbp]
\centerline{\includegraphics[width=\linewidth,height=8cm,width=12cm]{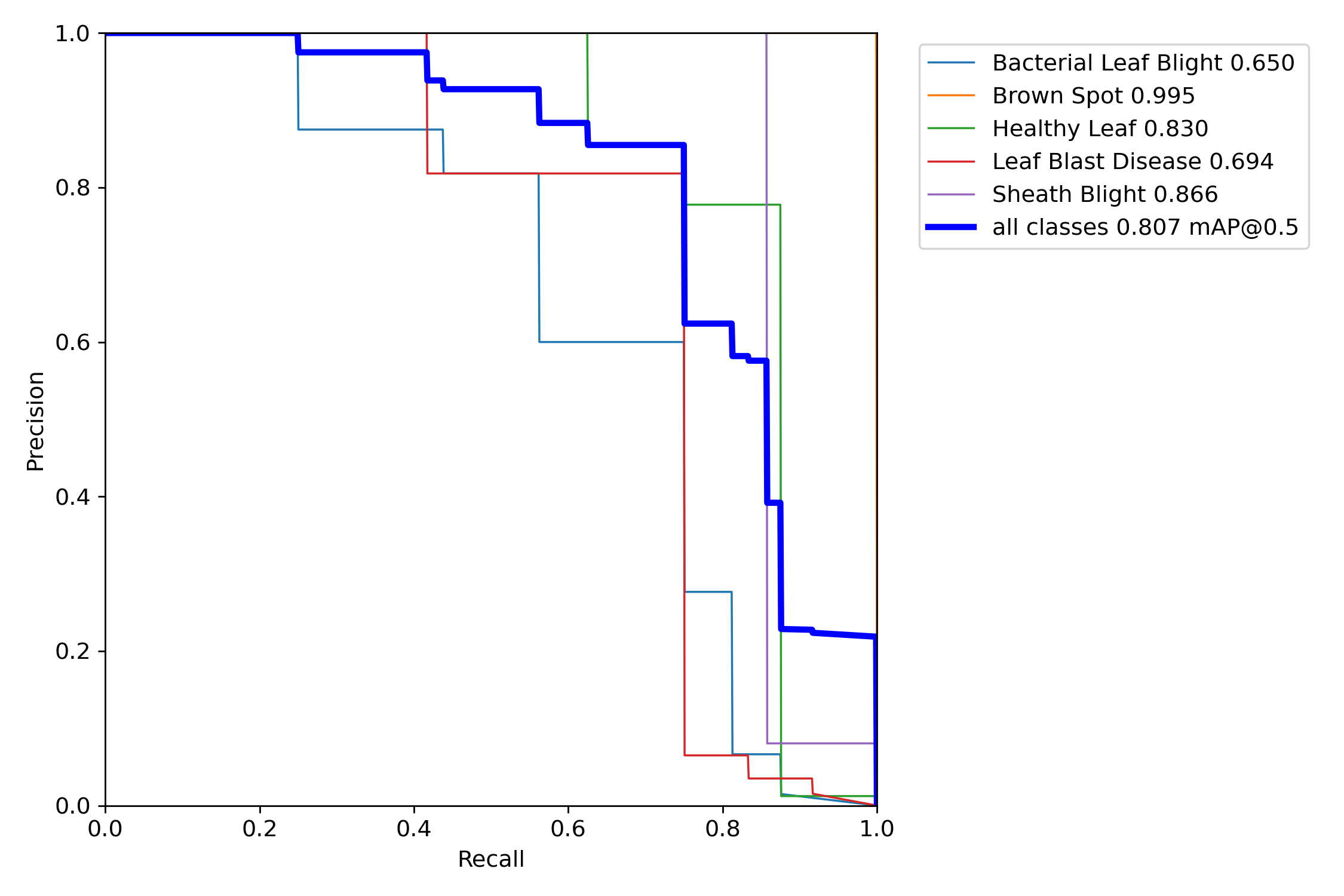}}
\caption{PR Curve}
\label{fig19}
\end{figure}
The compromise between precision and recall for various thresholds is depicted by the precision-recall curve. High precision is correlated with a low false positive rate, while big recall is correlated with a low false negative rate. A high area under the curve denotes both high recall and high precision. We discovered 0.807 mAP (Mean Average Precision) using the precision-recall curve, as depicted in Fig. 19.
\begin{figure}[htbp]
\centerline{\includegraphics[width=\linewidth,height=12cm,width=12cm]{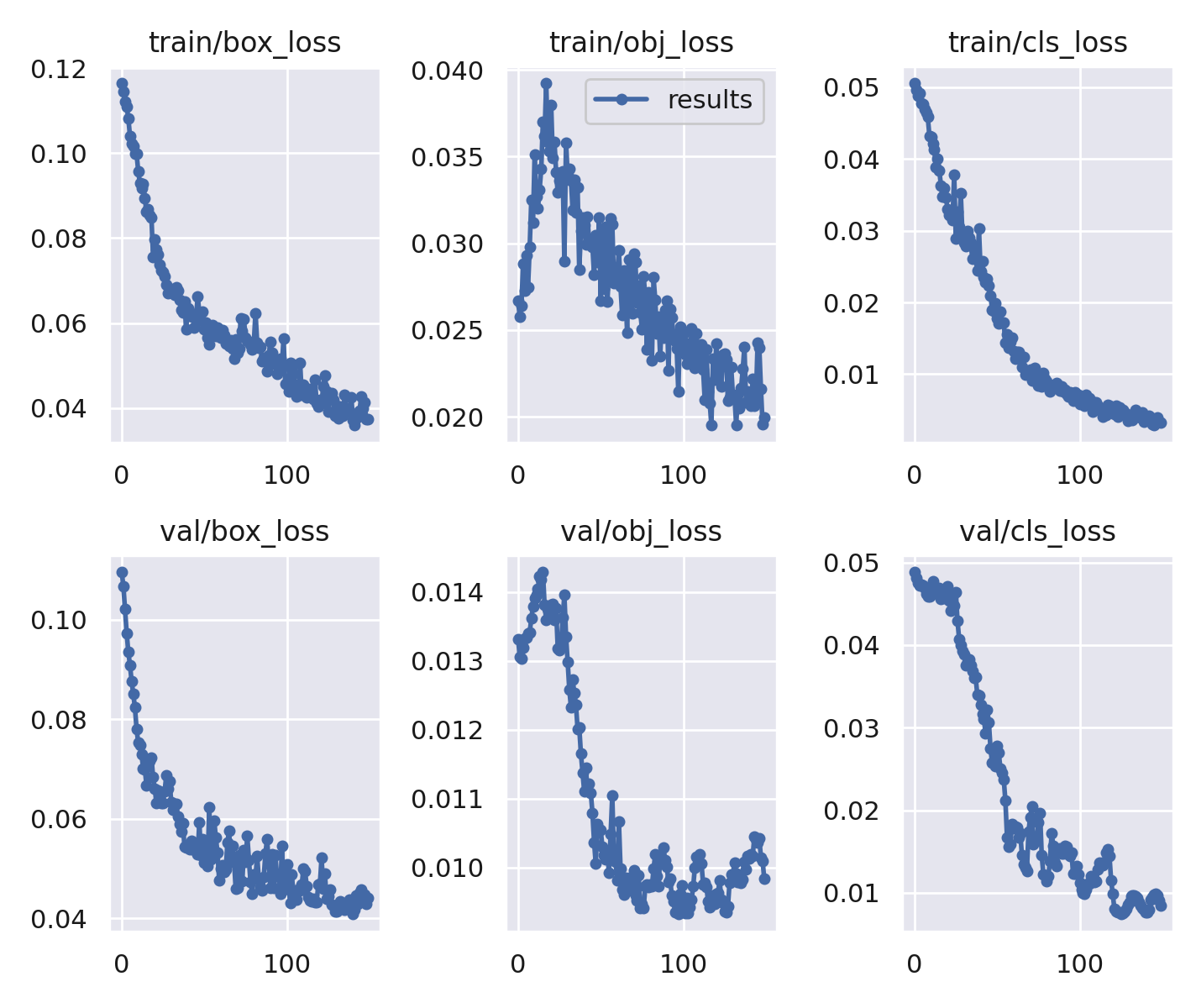}}
\caption{Dataset loss Graph}
\label{fig20}
\end{figure}
Using the training and verification sets, the network was trained. After 80 training batches, the detection frame loss, detection object loss, and classification loss value curves for the training and verification sets were established shown in Fig. 20.
\begin{figure}[htbp]
\centerline{\includegraphics[width=\linewidth,height=12cm,width=12cm]{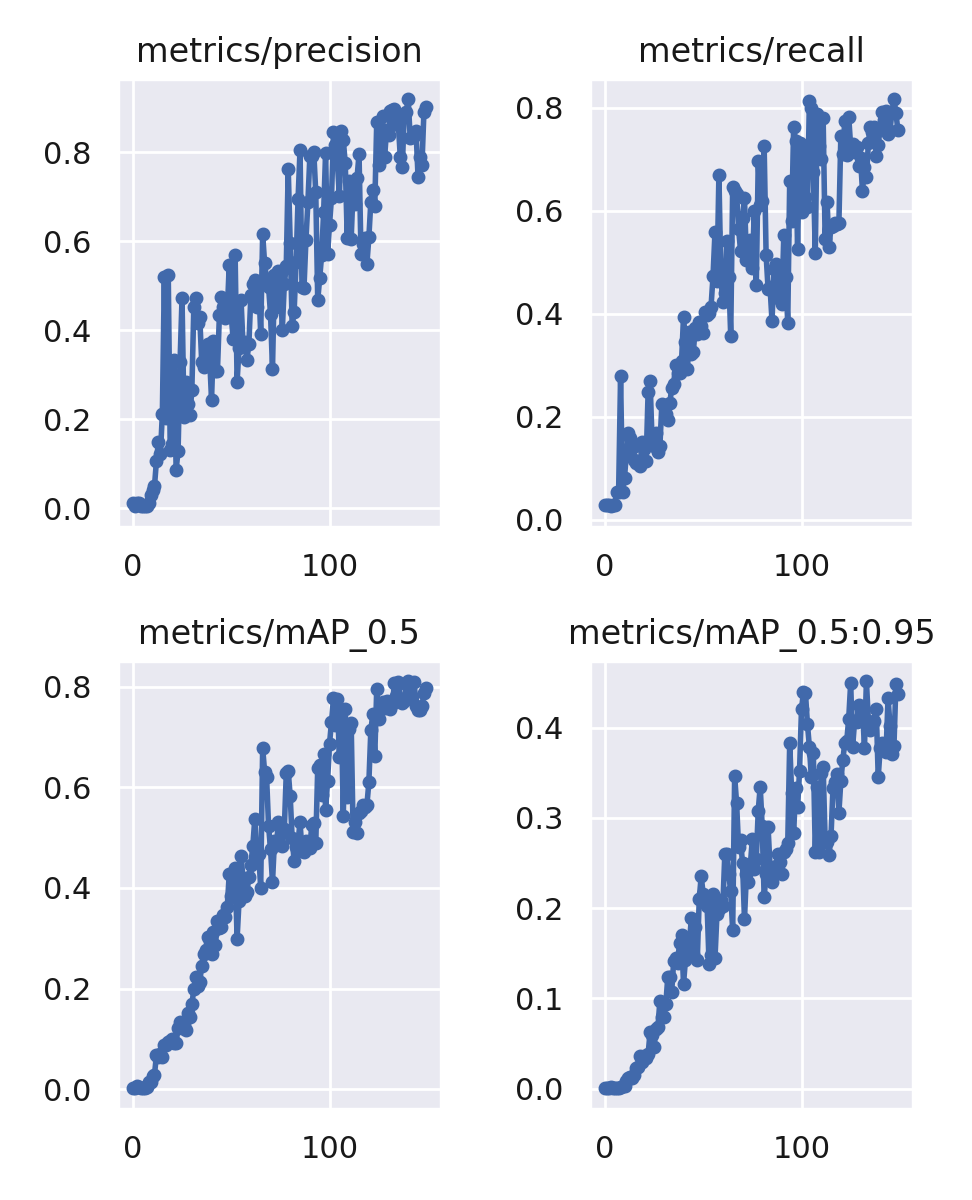}}
\caption{Result Graph}
\label{fig21}
\end{figure}
\begin{table}
  \centering
  \begin{tabular}{lllll}
    \cmidrule(r){1-5}
    Model     & F1   &Precision   & Recall   & mAP/AP\\
    \midrule
    YOLOv5    & 76   & 90         & 67       & 81  \\
    \bottomrule
  \end{tabular}
  \\
  \caption{Model Result Table}
  \label{tab:table}
\end{table}
\section{Conclusion}
Rice leaf disease has been extremely well detected using CNN of this recharge paper. The YOLO V5 is significantly more accurate than earlier techniques (SVM, VGG16 \& ResNet V2 ) etc. This image shows numerous rice leaf diseases that were identified using the YOLO V5 algorithm. A data set of approximately 1500 photographs was used in our research report. By using so many datasets during training, very good accuracy has been attained. In essence, the main illnesses that affect rice leaves have been addressed here. Thus, it may be argued that our effort will more significantly improve agriculture. CNN will be investigated in future study to classify other varieties of rice diseases and other plant leaf diseases.
\\

\end{document}